\documentclass[runningheads]{llncs}

\usepackage{times}
\usepackage{microtype}
\usepackage{graphicx}
\usepackage{amssymb}
\usepackage{amsmath}
\usepackage{tikz-cd}
\usepackage{dsfont}
\usepackage{mathrsfs}
\usepackage{makecell}
\usepackage{mathtools}
\usepackage{algorithm}
\usepackage{algorithmic}
\usepackage[T1]{fontenc}
\usepackage{appendix}
\usepackage{cite}
\usepackage{stmaryrd}
\usepackage[hidelinks]{hyperref}
\usepackage{orcidlink}

\newcommand{\dentails}{\mid\hskip-0.40ex\approx}
\newcommand{\ndentails}{\not\mid\hskip-0.40ex\approx}
\renewcommand{\bar}[1]{\overline{#1}}
\newcommand{\twiddle}{\mathrel|\joinrel\sim}

\DeclareSymbolFont{symbolsC}{U}{txsyc}{m}{n}
\DeclareMathSymbol{\strictif}{\mathrel}{symbolsC}{74}

\title{Towards Propositional KLM-Style Defeasible\\ Standpoint Logics}
\author{Nicholas Leisegang\inst{1}\orcidlink{0000-0002-8436-552X}\and
Thomas Meyer\inst{1}\orcidlink{0000-0003-2204-6969}\and
Sebastian Rudolph\inst{2,3}\orcidlink{0000-0002-1609-2080}}
\date{July 2024}

\authorrunning{N. Leisegang et al.}

\institute{University of Cape Town and CAIR, Cape Town, South Africa\\ 
\email{lsgnic001@myuct.ac.za}, \email{tommie.meyer@uct.ac.za}
\and
Technische Universität Dresden, Dresden, Germany\\
\and
Center for Scalable Data Analytics and Artificial Intelligence Dresden/Leipzig\\
\email{sebastian.rudolph@tu-dresden.de}}

\begin{document}
\maketitle
\begin{abstract}
The KLM approach to defeasible reasoning introduces a weakened form of implication into classical logic. This allows one to incorporate exceptions to general rules into a logical system, and for old conclusions to be withdrawn upon learning new contradictory information. Standpoint logics are a group of logics, introduced to the field of Knowledge Representation in the last 5 years, which allow for multiple viewpoints to be integrated into the same ontology, even when certain viewpoints may hold contradicting beliefs. In this paper, we aim to integrate standpoints into KLM propositional logic in a restricted setting. We introduce the logical system of Defeasible Restricted Standpoint Logic (DRSL) and define its syntax and semantics. Specifically, we integrate ranked interpretations and standpoint structures, which provide the semantics for propositional KLM and propositional standpoint logic respectively, in order to introduce ranked standpoint structures for DRSL. Moreover, we extend the non-monotonic entailment relation of rational closure from the propositional KLM case to the DRSL case. The main contribution of this paper is to characterize rational closure for DRSL both algorithmically and semantically, showing that rational closure can be characterized through a single representative ranked standpoint structure. Finally, we conclude that the semantic and algorithmic characterizations of rational closure are equivalent, and that entailment-checking for DRSL under rational closure is in the same complexity class as entailment-checking for propositional KLM.

\end{abstract}

\section{Introduction}

Within the field of symbolic AI, much work has been dedicated to reasoning with information which is incomplete, or is seemingly contradictory. One avenue in which contrasting beliefs is explored is through non-monotonic reasoning, which considers cases where conclusions can be made based on a given set of beliefs, and then withdrawn if new information arises that contradicts previous conclusions. A specific form of such non-monotonic reasoning was established by Kraus et al. \cite{kraus:nonmonotonic}, who introduced a defeasible implication $\twiddle$ into classical propositional logic, where the term $\alpha \twiddle \beta$ reads as ``$\alpha$ \textit{typically implies} $\beta$''. More specifically, this reads as saying that all the most typical occurrences of $\alpha$ should also satisfy $\beta$. Thus, we are able to incorporate rules with exceptions into our logic, since we can have instances where $\alpha$ is true but $\beta$ is not, and merely interpret this as some exceptional instance of $\alpha$. This approach to defeasible reasoning is referred to as the KLM framework.

Another instance where contradictory information occurs in knowledge representation, is where we wish to integrate the views of two agents into a single ontology or knowledge base, where the views of these agents may contradict each other. One way to do this is to weaken or exclude the contradicting beliefs between agents. However, this method may sacrifice the accuracy of the beliefs represented.  In order to remedy this, Gómez Álvarez and Rudolph \cite{alvarezrudolph:propositionalstdpt} introduce \textit{standpoint} modal operators, $\Box_s$ and $\Diamond_s$,  to the propositional case where $\Box_s \phi$ reads that ``\textit{it is unequivocal to $s$ that $\phi$}'' and $\Diamond_s \phi$ reads that ``\textit{it is possible to $s$ that $\phi$}''. Standpoint modal operators have also been introduced by Gómez Álvarez et al. \cite{alvarez:stdptlogicfocase} in the case of first-order and description logics . However, the above cases of standpoint logics continue to be monotonic in nature.

In this paper, we consider the case where we combine both approaches in order to represent, in a single logical framework, multiple agents who hold (possibly contradicting) beliefs, where these beliefs may be defeasible in their nature. We will also build a non-monotonic notion of entailment, which extends the well-known system of rational closure in propositional KLM \cite{lehmann:conditionalentail}. A motivating example for this case is given below.

\begin{example}\label{example:originaltomatoes}
    In the 19th century, crops imported into the USA were divided into fruits and vegetables, where fruits were exempt from import tax. This led to a court case on whether a tomato should be legally classified as a fruit or a vegetable. From a botanical standpoint, tomatoes are fruits and furthermore all fruits are also vegetables, since vegetables are considered all commonly eaten plants. This can be expressed using standpoint logics with the set:
    \[\{\Box_{B}(tomato \rightarrow fruit), \Box_{B}(fruit \rightarrow vegetable)\},\]
    where $B$ represents the botanical standpoint. However, the court considered a different standpoint based on the culinary use of tomatoes. This states that vegetables are those crops which are savoury and fruits are those which are sweet. That is,
    \[\{\Box_{C}(savoury \leftrightarrow vegetable), \Box_C(sweet \leftrightarrow fruit),\Box_{C}(tomato\twiddle savoury),\] 
    \[ \Box_{C}(fruit\twiddle \neg vegetable)\wedge \Box_{C}(vegetable\twiddle \neg fruit)\},\]
    where $C$ represents the ``culinary-use'' standpoint on crops. The first two statements above tell us the specified definitions for fruits and vegetables. The third statement tells us that tomatoes are usually used in savoury dishes, and the last proposition states that in a culinary perspective, fruits and vegetables are usually distinct from each other. The court concluded that from the ``culinary-use'' standpoint, tomatoes were usually considered vegetables, and therefore should be taxed. We can represent this by $\{L\preceq C, \Box_{L}(vegetable\rightarrow \neg fruit)\}$, where $L\preceq C$ tells us that the legal standpoint $L$ holds true each conclusion of $C$'s standpoint, and the second proposition states that fruits and vegetables are strictly distinct, legally speaking. Importantly, this system allows for both internal exceptions, and strict disagreements between standpoints. For example, $B$ holds that every fruit is a vegetable, while $L$ holds that no fruit is a vegetable. We also see that it is possible that a certain kind of exceptional tomato is not savoury from a culinary standpoint. In fact, from a culinary perspective it seems possible to consider a tomato a fruit and a vegetable (i.e. $\Diamond_C(tomato\rightarrow(fruit \wedge vegetable))$), although this is not possible from a legal perspective.
    
    \end{example}
    In order to formally analyse examples as the one above, we formally introduce \textit{Defeasible Restricted Standpoint Logic} (DRSL) and describe how this can be used to reason non-monotonically with multi-perspective knowledge bases. In Section 2, we provide a background on propositional KLM defeasibility, the non-monotonic entailment relation of rational closure and propositional standpoint logic in the classical case. In Section 3, we describe the syntax and semantics for DRSL. In Section 4, we describe how rational closure can be lifted from the propositional case to the case of standpoint logic, and show a correspondence between an algorithmic and a semantic definition of rational closure. Furthermore, we show that the algorithm for entailment-checking in rational closure is in the same complexity class as entailment-checking in the propositional case.
    
\section{Background}
\subsection{Propositional KLM and Rational Closure}

In this section we define the syntax and semantics for KLM in the propositional case, as well as recall certain results which we rely upon later on. In particular, we are considering the case where $\twiddle$ forms a rational consequence relation as is defined by Lehmann and Magidor \cite{lehmann:conditionalentail}. We use a slightly more expressive syntax than the traditional KLM which allows for conjunctions between propositions.

    That is, we define the language of KLM propositional logic $\mathcal{L}^{\twiddle}$, over a set of propositional atoms $\mathcal{P}$, as defined by
    \[\phi::= \alpha\mid \alpha\twiddle \beta\mid \phi\wedge \phi,\]

where $\alpha$ and $\beta$ are Boolean formulas with atoms in $\mathcal{P}$.

The semantics for $\mathcal{L}^{\twiddle}$ is referred to as ranked interpretations. These were defined originally as a Kripke-style semantics \cite{kraus:nonmonotonic}. However, in this paper we use an equivalent notion of ranking functions to define such an interpretation \cite{casini:beyondratclosure}. For the definition below $\mathcal{U}$ refers to the set of classical interpretations for the atoms in $\mathcal{P}$. That is, each $u\in\mathcal{U}$ is a map that assigns each $p\in\mathcal{P}$ a truth value.

A \textbf{ranked interpretation} is therefore defined as a function $R:\mathcal{U}\rightarrow \mathds{N}\cup \{\infty\}$, such that the following convexity property is satisfied: if $R(u)<\infty$, then for every $0\leq j<R(u)$ there exists $v\in \mathcal{U}$ such that $R(v)=j$.

A  ranked interpretation intuitively tells us how typical a state of the world ought to be, with those valuations with higher ranks being less typical than those with lower ranks. Those valuations with rank $\infty$ then refer to states of the world that are ``impossible''. We define $\llbracket \alpha \rrbracket=\{u\in\mathcal{U}\mid u\Vdash \alpha\}$ for any Boolean formula $\alpha$, and let $\mathcal{U}^R=\{u\in \mathcal{U}\mid R(u)\neq \infty\}$. Then we say $R\Vdash \alpha$ iff $\llbracket \alpha \rrbracket\supseteq\mathcal{U}^R$. We also say that $R\Vdash \alpha\twiddle \beta$ if and only if $\{u\in \llbracket\alpha\rrbracket\mid R(v)\nless R(u)\text{ for all } v\in\llbracket\alpha\rrbracket\}\subseteq \llbracket\beta\rrbracket$. That is, if all the valuations satisfying $\alpha$ with a minimal rank also satisfy $\beta$, or ``the most typical $\alpha$-instances are instances of $\beta$''. We further note that a Boolean formula $\alpha$ can be expressed equivalently by a defeasible implication $\neg\alpha\twiddle\bot$ \cite{kraus:nonmonotonic}.

For a set $\mathcal{K}\in\mathcal{L}^{\twiddle}$ we also say that $R\Vdash \mathcal{K}$ when $R\Vdash \phi$ for all $\phi\in\mathcal{K}$. We therefore define that $R\Vdash \phi_1\wedge \phi_2$ if and only if $R\Vdash \{\phi_1,\phi_2\}$. The reason we include conjunctions in the language $\mathcal{L}^{\twiddle}$ in this paper is to allow for increased expressivity when we bound formulas by standpoint modal operators in Section \ref{section:DRSLsyntaxsemantic}.

\begin{example}
    We consider a knowledge base $\mathcal{K}=\{p\rightarrow b,b\twiddle f,p\twiddle \neg f\}$. Then a ranked interpretation which is a model of $\mathcal{K}$ is represented in the table below, where each valuation is placed in the row corresponding to its rank. Moreover, each valuation is represented by a sequence of atomic propositions in $\mathcal{K}$, where a bar is placed over the top of each atom which is false in the valuation. For example, the valuation $pb\bar{f}$ is the valuation where $p$ and $b$ are true and $f$ is false, and has rank $1$ in the table below:
    
\centering \begin{tabular}{|c|c |}
\hline
$\infty$ & $p\bar{b}f,p\bar{bf}$\\
\hline
$2$& $pbf$\\
\hline
$1$ & $pb\bar{f}, \bar{p}b\bar{f}$\\
\hline
$0$&$\bar{p}bf,\bar{pb}f,\bar{pbf}$\\
\hline
\end{tabular}
\end{example}

It is also well-known in the KLM literature that there are multiple non-equivalent notions of defeasible entailment defined to characterize what a KLM knowledge base $\mathcal{K}\subseteq \mathcal{L}^{\twiddle}$ entails. The first of these, defined by Lehmann and Magidor \cite{lehmann:conditionalentail}, is \textit{rational closure}. This is defined using a single ranked interpretation $R^{\mathcal{K}}_{RC}$, or equivalently, via an algorithmic approach. The algorithmic definitions are given by the \texttt{RCProp} algorithm for entailment-checking in Algorithm \ref{algorithm:rcprop} which in turn calls on the \texttt{BaseRank} algorithm described in Algorithm \ref{algorithm:baserank}.  These are originally proposed by Freund \cite{freund:preferentilreasoning} but resemble closer those occurring in Casini et al. \cite{casini:beyondratclosure}. In these algorithms, we assume that each knowledge base $\mathcal{K}$ only contains elements of the form $\alpha\twiddle \beta$, and that entailment checking is only done for elements of this form. This can be done by changing each Boolean formula $\alpha$ into $\neg\alpha\twiddle\bot$ and by splitting any conjunction of formulas into a set of its conjuncts.

In order to define the semantic notion of rational closure, one uses an order on models of $K\subseteq\mathcal{L}^{\twiddle}$ denoted $\preceq_{\mathcal{K}}$, where $R_1\preceq_{\mathcal{K}} R_2$ iff $R_1(u)\leq R_2(u)$ for all $u\in \mathcal{U}$.\footnote{It is worth noting this holds with respect to valuations $\mathcal{U}$ on $\mathcal{P}$, although not all atomic propositions in $\mathcal{P}$ may occur in $\mathcal{K}$. In the literature these two sets of atoms are often considered to be the same, but this is not the case when we introduce standpoints, since there may be atoms that do not occur in the beliefs of one standpoint that occur in the beliefs of another.} It is observed by Giordano et al. \cite{giordano:semanticRC} that the set of ranked models of $\mathcal{K}$, denoted $\mathcal{R}^\mathcal{K}$, has a minimum with respect to the order $\preceq_{\mathcal{K}}$. This minimal element is denoted $R^{\mathcal{K}}_{RC}$. This assumption states that we ought to reason as though things are as typical as they can be unless we have reason to believe otherwise. This corresponds to moving valuations in $\mathcal{U}$ to their lowest rank possible without contradicting anything occurring in the knowledge base $\mathcal{K}$. From this, we obtain the following representation result for propositional rational closure. For a knowledge base $\mathcal{K}$ and a defeasible implication $\alpha\twiddle \beta$,  $\text{RCProp}(\mathcal{K},\alpha\twiddle \beta)$=True if and only if $R^{\mathcal{K}}_{RC}\Vdash \alpha\twiddle \beta$ \cite{lehmann:conditionalentail, Pearl:SystemZ}. This result can be extended to conjunctions of defeasible implications by once again treating them as sets of their conjuncts.

\begin{figure}[t]
\noindent\begin{minipage}[t]{0.49\textwidth}
\begin{algorithm}[H]
			\caption{BaseRank}
			\label{algorithm:baserank}
			\textbf{Input}: A propositional KLM knowledge base $\mathcal{K}$\\
			\mbox{\textbf{Output}: An ordered tuple} \mbox{$(R_0,...,R_{n-1},R_\infty,n)$}\\[-2.5ex] 
			\begin{algorithmic}[1]
			\STATE $i:=0$;
            \STATE $E_0:=\{\alpha\rightarrow\beta\mid \alpha\twiddle\beta \in\mathcal{K}\}$;
            \REPEAT
            \STATE \mbox{$E_{i+1}:=\{\alpha\rightarrow\beta\in E_i\mid E_i\vDash\neg \alpha\}$;}
            \STATE $R_i:=E_i-E_{i+1}$;
            \STATE $i:=i+1$;
            \UNTIL{$E_{i-1}=E_i$;}
            \STATE $R_{\infty}:=E_{i-1};$
            \IF{$E_{i-1}=\emptyset$}
            \STATE $n:=i-1$;
            \ELSE
            \STATE $n:=i$;
            \ENDIF
            \RETURN $(R_0,...,R_{n-1},R_\infty,n)$
		\end{algorithmic}
		\end{algorithm}
  \end{minipage}
  \hfill
  \begin{minipage}[t]{0.475\textwidth}
\begin{algorithm}[H]
			\caption{RCProp}
			\label{algorithm:rcprop}
			\textbf{Input}: A propositional KLM knowledge base $\mathcal{K}$ and a defeasible implication $\alpha\twiddle\beta$\\
			\textbf{Output}: True if $\mathcal{K}\dentails_{RC}\alpha\twiddle\beta$ and False otherwise
			\begin{algorithmic}[1]
			\STATE\mbox{$(R_0,\!...,R_{n-1},R_\infty,n):=\text{BaseRank}(\mathcal{K})$;}
            \STATE $i:=0$;
            \STATE $R:=\bigcup^{j<n}_{j=0}R_j$;
            \WHILE{$R_\infty\cup R\vDash \neg \alpha$ and $R\neq \emptyset$}
            \STATE $R:=R-R_i$;
            \STATE $i:=i+1$;
            \ENDWHILE
            \RETURN $R_\infty\cup R\vDash \alpha\rightarrow\beta$
		\end{algorithmic}
		\end{algorithm}
\end{minipage}
\end{figure}

\begin{definition}
    Consider $\phi\in\mathcal{L}^{\twiddle}$. If $\phi=\alpha\twiddle\beta$, we say $\phi$ is in the rational closure of a knowledge base $\mathcal{K}$, or write $\mathcal{K}\dentails_{RC}\phi$, when $R^{\mathcal{K}}_{RC}\Vdash \alpha\twiddle \beta$. Equivalently, where $\text{RCProp}(\mathcal{K},\alpha\twiddle \beta)$=True. If $\phi=\phi_1\wedge \phi_2$, we say $\mathcal{K}\dentails_{RC}\phi$ when $\mathcal{K}\dentails_{RC}\phi_1$ and $\mathcal{K}\dentails_{RC}\phi_2$.
\end{definition}

We also note the original complexity result for rational closure in the propositional case, which states that \texttt{RCProp} is computable in $\text{P}^{\text{NP}}$ \cite{lehmann:conditionalentail}.

\subsection{Standpoint Logics}

Standpoint logic is introduced by Gómez Álvarez and Rudolph \cite{alvarezrudolph:propositionalstdpt} in the propositional case using the following syntax and semantics.

\begin{definition}
    Consider a vocabulary $\mathcal{V}=(\mathcal{P},\mathcal{S})$, where $\mathcal{P}$ is a finite set of propositional atoms and $\mathcal{S}$ is a finite set of standpoint symbols containing the universal standpoint~$*$. The language $\mathcal{L}_{\mathds{S}}$ over $\mathcal{V}$ is defined by
    $\phi::=s_1\preceq s_2\mid p\mid \neg \phi\mid \phi \wedge \phi\mid \Box_s \phi,$
   where $s_1,s_2,s\in\mathcal{S}$ and $p\in \mathcal{P}$. 
\end{definition}

In this syntax, $s_1\preceq s_2$ are called \textit{standpoint sharpening statements}. The Boolean connectives $\vee$, $\rightarrow$, $\leftrightarrow$ are defined via $\neg$ and $\wedge$ in their usual manner, and for each standpoint $s\in\mathcal{S}$ we define $\Diamond_s:=\neg\Box_s\neg$. The semantics for $\mathcal{L}_{\mathds{S}}$ are defined as follows.

\begin{definition}
    A \textbf{standpoint structure} is a triple $M=(\Pi,\sigma,\gamma)$ where $\Pi$ is a non-empty set of precisifications, $\sigma:\mathcal{S}\rightarrow P(\Pi)$ is a function which assigns each standpoint a non-empty set of precisifications and $\sigma(*)=\Pi$, and $\gamma:\Pi\rightarrow P(\mathcal{P})$ is a function which assigns each precisification a set of atoms. 
\end{definition}

Intuitively, the map $\sigma$ allows one to allocate to a standpoint $s$ the set of all ``reasonable ways to make $s$'s beliefs correct'', and $\gamma$ assigns a set of basic propositions which are ``true'' in that precisification. Equivalently, it associates to each $\pi$ a classical valuation on $\mathcal{P}$, describing the contents of what the precisification $\pi$ holds true. This notion is emphasised by the definition of the satisfaction relation $\Vdash$.

\begin{definition}
    For a standpoint structure $M$ and a precisification $\pi\in\Pi$, we define the satisfaction relation $\Vdash$ inductively as follows:
    \begin{itemize}
        \item $M,\pi\Vdash p$ iff $p\in\gamma(\pi)$.
        \item $M,\pi\Vdash \neg \phi$ iff $M,\pi\nVdash \phi$.
         \item $M,\pi\Vdash \phi_1\wedge \phi_2$ iff $M,\pi\Vdash \phi_1$ and $M,\pi\Vdash \phi_2$.
         \item $M,\pi\Vdash \Box_s\phi$ iff $M,\pi'\Vdash \phi$ for all $\pi'\in \sigma(s)$.
         \item $M,\pi\Vdash s_1\preceq s_2$ iff $\sigma(s_1)\subseteq \sigma(s_2)$
         \item $M\Vdash\phi$ iff $M,\pi\Vdash \phi$ for all $\pi\in \Pi$.
    \end{itemize}
    In the above $\phi,\phi_1\phi_2\in\mathcal{L}_{\mathds{S}}$, $s,s_1,s_2\in\mathcal{S}$ and $p\in \mathcal{P}$.
\end{definition}

Based on these semantics, and those for KLM, we propose an integration of these two logics in the following section.

\section{Defeasible Restricted Standpoint Logic}\label{section:DRSLsyntaxsemantic}

	\subsection{Syntax}
 In the language of DRSL, we consider the case where standpoint modal operators are applied to propositional KLM formulas, with some restrictions of how standpoint bound KLM formulas can be combined. Formally, this is defined below.
  
		\begin{definition}
			A \textbf{vocabulary} is a pair $\mathcal{V}=(\mathcal{P},\mathcal{S})$ where $\mathcal{P}$ is a finite set of propositional atoms and $\mathcal{S}$ is a finite set of standpoints containing the universal standpoint $*$. The language $\mathcal{L}^{\twiddle}_\mathds{S}$ is defined as follows,
            \[\psi::=\phi\mid \#_s \psi \mid \psi\wedge \psi\quad\text{ or }\quad e::=s_1\preceq s_2,\]
			where $\#=\Box$ or $\#=\Diamond$, $s_1$ and $s_2$ are standpoints and $\phi$ is a formula in $\mathcal{L}^{\twiddle}$ with atoms in $\mathcal{P}$.
		\end{definition}
          
		Note here that we define the syntax in such a way that we can only apply standpoint modal operators to the outside of existing formulas in the language of KLM logic. For example, $\Box_s (p\twiddle (q\vee r))$ would be a valid formula in $\mathcal{L}^{\twiddle}_\mathds{S}$ while $\Box_s p\twiddle \Box_{s'} (q\vee r)$ is not valid. We also define the dual operator $\Diamond_s$ explicitly since we do not allow for negation of modal operator bound formulas. We also do not allow for disjunctions of formulas in our syntax. Due to this, we lose the ability to express the related operator $\mathcal{D}_s$ defined by $\mathcal{D}_s\phi=\Box_s\neg \phi\vee \Box_s \phi$, which is read to mean that $\phi$'s truth is determined in $s$'s standpoint \cite{alvarezrudolph:propositionalstdpt}. However, we are able to express in the case where $\phi$ is Boolean the additional standpoint operator $\mathcal{I}_s\phi:= \Diamond_s\neg \phi \wedge \Diamond_s\phi$, which states that $\phi$'s truth is not determined in $s$'s standpoints. If $\phi$ is not Boolean, and contains some defeasible implication, we are not able to express this using our given syntax.\\

	\subsection{Semantics}
		The semantics we introduce to this standpoint logic follows the intuition given by Gómez Álvarez et al. \cite{alvarez:stdptlogicfocase} such that we define a set of ``precisifications'' of formulas, which correspond to ``worlds'' in a possible-worlds style semantics. Moreover, they intuitively refer to making precise those views held within a standpoint, and different ways in which such a standpoint can be made precise. Every precisificaton is then mapped to an underlying semantic structure for the ``base logic''. In first order, or description logics, this corresponds to a first-order interpretation structure \cite{alvarez:stdptlogicfocase}. Using the standard semantics for KLM defeasible logic, we will construct a ranked interpretation for each given precisification.\\
  
		\begin{definition}
			Given a vocabulary $\mathcal{V}=(\mathcal{P},\mathcal{S})$, a \textbf{ranked standpoint structure} is a triple $M=(\Pi, \sigma, \gamma)$ where:
			
			\begin{itemize}
				\item $\Pi$ is a non-empty set of precisifications.
				\item $\sigma:\mathcal{S}\rightarrow P(\Pi)$ is a function that maps each standpoint symbol to a non-empty set of precisifications and in particular, $\sigma(*)=\Pi$.
				\item $\gamma:\Pi \rightarrow \mathcal{R}$ is a function, where $\mathcal{R}$ is the set of ranked interpretations over $\mathcal{P}$. That is, for each precisification $\pi$, $\gamma(\pi)$ is a ranked interpretation.
			\end{itemize}
		\end{definition}
		
		We say that $M$ is \textit{valid} if and only if for each $s\in\mathcal{S}$, for every $\pi\in\sigma(s)$, there exists some valuation $u$ such that $\gamma(\pi)(u)\neq \infty$.   We then say that a ranked standpoint structure is a \textit{model} for a formula $\xi\in \mathcal{L}^{\twiddle}_{\mathds{S}}$ if and only if $M\Vdash \xi$.\footnote{This definition of a model extends to sets of formulas in the usual way.} The relation $\Vdash$ is defined as follows.
		
		\begin{definition}
			Given a ranked interpretation structure $M$ and a precisification $\pi\in \Pi$, the satisfaction relation $\Vdash$ is defined inductively as follows:
			\begin{itemize}
				\item $M, \pi \Vdash \phi$ iff $\gamma(\pi)\Vdash \phi$ for the ranked interpretation $\gamma(\pi)$.
				\item $M, \pi \Vdash \Box_s \psi$ iff $M,\pi' \Vdash \psi$ for all $\pi'\in \sigma(s)$.
				\item $M, \pi \Vdash \Diamond_s \psi$ iff $M,\pi' \Vdash \psi$ for some $\pi'\in \sigma(s)$.
                \item $M,\pi\Vdash \psi_1\wedge \psi_2$ iff $M,\pi\Vdash \psi_1$ and $M,\pi\Vdash \psi_2$.
                \item$M,\pi\Vdash s_1\preceq s_2$ iff $\sigma(s_1)\subseteq \sigma (s_2)$.
                \item $M\Vdash \psi$ iff $M,\pi\Vdash \psi$ for all $\pi\in\Pi$.
			\end{itemize} 
 where $\phi\in\mathcal{L}^{\twiddle}$, $\psi\in\mathcal{L}^{\twiddle}_{\mathds{S}}$ and $s,s_1,s_2\in\mathcal{S}$.
\end{definition}
		
			Something to note on this semantics is that the application of multiple modal operators over a single formula reduces to just one formula, as is in the case for other standpoint logics \cite{alvarezrudolph:propositionalstdpt}. That is, $M\Vdash \#^1_{s_1}\#^2_{s_2}...\#^n_{s_n}\phi$ if and only if $M\Vdash \#^n_{s_n}\phi$, for any ranked standpoint structure $M$, any $s_1,...,s_n\in\mathcal{S}$ and any $\phi\in\mathcal{L}^{\twiddle}$. Therefore, in the rest of this paper we only consider those formulas bound by one or no standpoint modal operators. 

            Another result which is evident from these semantics is that for any formula $\phi\in \mathcal{L}^{\twiddle}$ and any ranked standpoint structure $M$, we have $M\Vdash \phi$ if and only if $M\Vdash \Box_* \phi$. This is also the case in other versions of standpoint logics \cite{alvarezrudolph:propositionalstdpt}. These two facts, along with the following Lemmas allow us to define a \textit{normal form} for a formula in DRSL.

            \begin{lemma}\label{lemma:boxdistribution}
                For any ranked standpoint structure $M$ and any $\psi_1,\psi_2\in \mathcal{L}^{\twiddle}_{\mathds{S}}$, we have that $M\Vdash \Box_s(\psi_1\wedge \psi_2)$ if and only if $M\Vdash \Box_s\psi_2\wedge \Box_s\psi_2$.
            \end{lemma}

            \begin{lemma}\label{lemma:diamonddistribution}
            Let $M$ be a ranked standpoint structure, $\phi_1,...,\phi_k\in\mathcal{L}^{\twiddle}$. Then $M\Vdash \Diamond_s(\phi_1\wedge\#^2_{s_2}\phi_2\wedge ...\wedge\#^k_{s_k} \phi_k)$ if and only if $M\Vdash\Diamond_s\phi_1\wedge\#^2_{s_2}\phi_2\wedge ...\wedge\#^k_{s_k} \phi_k$.
            \end{lemma}

            From these two results we obtain the following.

            \begin{corollary}\label{corollary:propositionnormalform}
                Any DRSL formula $\psi$ can be represented equivalently by a formula in the form $\#^1_{s_1}\phi_1\wedge...\wedge \#^k_{s_k}\phi_k$, where $\phi_1,...,\phi_n\in\mathcal{L}^{\twiddle}$.
            \end{corollary}

            For example, the formula \[\psi=\Diamond_s((p_1\twiddle p_2)\wedge(p_3\twiddle p_4)\wedge\Box_t(p_5\twiddle p_6)\wedge (p_7\twiddle p_8)))\wedge p_9\] can be equivalently represented by 
            \[\widetilde{\psi}= \Diamond_s((p_1\twiddle p_2)\wedge(p_3\twiddle p_4))\wedge\Box_t(p_5\twiddle p_6)\wedge\Box_t (p_7\twiddle p_8)\wedge \Box_*p_9\]

            \begin{definition}
                Any formula of the form specified in Corollary \ref{corollary:propositionnormalform} is said to be in \textbf{normal form}. The process of rewriting a formula $\psi$ in normal form is called \textbf{normalization} of $\psi$.
            \end{definition}

        This definition for the semantics for DRSL gives rise to a natural Tarskian notion of entailment, which is analogous to \textit{ranked entailment} defined in the propositional case of KLM \cite{lehmann:conditionalentail}.

        \begin{definition}
             The \textbf{ranked entailment} relation $\vDash_R$ is defined as follows. Consider a knowledge base $\mathcal{K}$ of propositions in $\mathcal{L}^{\twiddle}_{\mathds{S}}$ and a proposition $\xi\in \mathcal{L}^{\twiddle}_{\mathds{S}}$. Then $\mathcal{K}\vDash_R \xi$ if and only if every ranked standpoint structure $M$ which is a model of $\mathcal{K}$ is also a model of $\xi$.
        \end{definition}

        Although ranked entailment is a notion worth investigating further, it is monotonic, and therefore does not give us the desired system of non-monotonic reason we wish to construct in the DRSL case. This is shown in the lemma below.

        \begin{lemma}\label{lemma:rankedentailmentmonotonic}
            Let $\mathcal{K}$ be a DRSL knowledge base. Then for any $\psi\in \mathcal{L}^{\twiddle}_{\mathds{S}}$, we have that $\mathcal{K}\vDash_R \xi$ implies $\mathcal{K}\cup\{\psi\}\vDash_R \xi$.
        \end{lemma}

        Therefore, in order to construct some kind of non-monotonic entailment for DSRL, we propose an extension of rational closure, as defined by Lehmann and Magidor \cite{lehmann:conditionalentail} in the propositional case, for standpoint logics.
        
        \section{Rational Closure for DRSL}\label{section:DRSLrationalclosure}
		In this section we define an algorithm for rational closure in DRSL. Furthermore, we show that this algorithm can also be described semantically via a single ranked standpoint structure. This is analogous to the semantic and algorithmic representations of rational closure in the propositional case as is described by Lehman and Magidor\cite{lehmann:conditionalentail}.

  \subsection{Rational Closure Algorithm}
  
		In this section we introduce the rational closure algorithm for DRSL, and describe its complexity. This begins by constructing a ``representative set'' of KLM knowledge bases for each standpoint $s\in\mathcal{S}$. We split a knowledge base with formulas in $\mathcal{L}_{\mathds{S}}^{\twiddle}$ into several knowledge bases in $\mathcal{L}^{\twiddle}$ which represent the KLM-knowledge bases that each standpoint $s$ takes into account. In the base case, we determines a knowledge based only on those formulas that are unequivocal for $s$, called $K_s$, which include formulas that are bound by $\Box_t$ where $t=s$ or $t$ is some standpoint for which ``subsumes'' $s$ in the sense that $s$ is a sharpening of $t$. Then we construct a separate knowledge base for the standpoint $s$ for each formula of the form $\Diamond_s\phi$ for $\phi\in\mathcal{L}^{\twiddle}$. That is, we create a new knowledge base representing the precisification of $s$ where $\phi$ is true. We then associate each standpoint with a set of representative KLM knowledge base, denoted $\text{Know}_s$, which includes $K_t$ and $K^\phi_t$ where $t=s$ or $t$ is a sharpening of $s$. This is described in Algorithm \ref{alg:standpointsplit}.
  
\begin{example}\label{example:tomatosplit}
      Using the same knowledge base in Example \ref{example:originaltomatoes} we can split our DRSL knowledge base into three KLM knowledge bases:
      \begin{align*}
        K_B=&\{tomato\rightarrow fruit, fruit\rightarrow vegetable\}&&\\
        K_C=&\{savoury \leftrightarrow vegetable, sweet \leftrightarrow fruit, tomato\twiddle savoury,\\ 
        &fruit\twiddle \neg vegetable, vegetable\twiddle \neg fruit\}&&\\
        K_L=&\{savoury \leftrightarrow vegetable, sweet \leftrightarrow fruit, tomato\twiddle savoury,\\ 
        &fruit\rightarrow\neg vegetable\}&&
    \end{align*}
    In $K_B$ and $K_C$ we simply include those propositions bound by $\Box_B$ and $\Box_C$, respectively. However, in the case of $K_L$, since $L\preceq C$ is in our knowledge base, any statement unequivocal to $C$ must also be unequivocal to $L$. Hence, we include statements bound by $\Box_L$ and $\Box_C$ in $K_L$. Then, we associate with each standpoint the sets of knowledge bases ``relevant'' to those standpoints, giving us:
    $\text{Know}_B=\{K_B\}$, $\text{Know}_C=\{K_C,K_L\}$ and  $\text{Know}_L=\{K_L\}$.Here note that since $L\preceq C$, we have to include $K_L$ in $\text{Know}_C$, since $L$ is a sharpening of $C$'s standpoint and therefore is seen as a more specific version of $C$'s view.    
    If we were to add a formula of the form $\Diamond_L\phi$ to our knowledge base, this would create another set in the splitting defined by $K_L^\phi=K_L\cup \{\phi\}$. Intuitively, this means if our knowledge base asserts that it is possible for $L$ to believe that $\phi$ is true, then we must consider the possibility of $L$'s view where this is true, but that it is a divergent possible precisification to the set of statements $K_L$ which are those unequivocal to $L$. In this case we would also obtain that $\text{Know}_C=\{K_C,K_L,K_L^\phi\}$ and  $\text{Know}_L=\{K_L,K_L^\phi\}$.
    \end{example}

  However, before we implement our splitting algorithm on a knowledge base $\mathcal{K}\subseteq \mathcal{L}_{\mathds{S}}^{\twiddle}$, we reduce $\mathcal{K}$ into a canonical form. We first \textit{normalize} each formula in $\mathcal{K}$ so that it only contains formulas in normal form. We then say that a DRSL knowledge base $\mathcal{K}$ is \textit{conjunction-free} if there are no formulas of the form $\psi_1\wedge \psi_2$ in $\mathcal{K}$. If $\mathcal{K}$ is not conjunction free, it is easy to observe that we can construct an equivalent knowledge base $\mathcal{K'}$ by splitting any conjunction in $\mathcal{K}$ into a set of its conjuncts.
  If $\mathcal{K}$ contains only normalized formulas and is conjunction free, we say that $\mathcal{K}$ is in \textit{normal form}. That is, it only contains formulas of the form $\#_s\phi$, where $\phi$ is in $\mathcal{L}^{\twiddle}$.

  We are then able to define Algorithm \ref{alg:standpointsplit}, which splits a DRSL knowledge base into a set of propositional KLM knowledge bases. We then apply propositional rational closure point-wise on these sets for entailment-checking over a DRSL knowledge base. A last definition used in the algorithm is as follows.

  \begin{definition}
      Let $\mathcal{K}$ be a DRSL knowledge base. For standpoint symbols $s$ and $t$, we say that $s\preceq^+ t$ if $t=s$, $t=*$ or there exists a finite sequence of standpoint symbols $t_1,...,t_n$ such that $s\preceq t_1,t_1\preceq t_2,...,t_{n-1}\preceq t_n,t_n\preceq t\in \mathcal{K}$.
  \end{definition}

        After splitting a DRSL knowledge base into several propositional KLM knowledge bases, we apply another algorithm for checking whether a (non-standpoint sharpening) formula $\xi\in\mathcal{L}^{\twiddle}_{\mathds{S}}$ is in the rational closure of a DRSL  knowledge base $\mathcal{K}$, and therefore whether $\mathcal{K}$ defeasibly entails $\xi$. In order to do this we presume the ability to call on Algorithm \ref{algorithm:rcprop}, for rational closure in the propositional case. We define this in Algorithm \ref{RCStandpoint}.

\begin{figure}[t]
\noindent\begin{minipage}[t]{0.48\textwidth}
		\begin{algorithm}[H]
			\caption{StandpointSplit}
			\label{alg:standpointsplit}
			\textbf{Input}: A defeasible standpoint knowledge base $\mathcal{K}$ in normal form.\\
			\textbf{Output}: A set of propositional KLM knowledge bases associated to each standpoint in $\mathcal{K}$,\\ denoted $\bigcup_{s\in S\cup\{*\}}\{\text{Know}_s\}$.\\[-2.5ex]
			\begin{algorithmic}[1] 
				\STATE Define $S$ as the set of standpoints occurring in the formulas of $\mathcal{K}$.
				\FOR{$s\in S$}
				\STATE $K_s:=\emptyset$;
				\WHILE{$t\in S$ and $s\preceq^+ t$}
				\FOR{$\Box_t\phi\in\mathcal{K}$}
				\STATE $K_s:=K_s\cup\{\phi\}$;
				\ENDFOR
				\ENDWHILE
                \ENDFOR
                \FOR{$s\in S$}
                \FOR{$\Diamond_s\phi\in \mathcal{K}$}
				\STATE$K_s^{\phi}:=K_s\cup\{\phi\}$;
				\ENDFOR
				\ENDFOR
                \FOR{$s\in S$}
                \STATE $S_{\preceq s}:=\{t\in S\mid t\preceq^{+} s\}$;
                \STATE $\text{Know}_s:=\cup_{t\in S_{\preceq s}}(\{K_t\}\cup$\\ \hspace{14mm}$\{K^\phi_t\mid \Diamond_t\phi\in\mathcal{K}\})$; 
                \ENDFOR
                \STATE $\text{Know}_*=\bigcup_{s\in S}\text{Know}_s-K_*$;
				\RETURN $\bigcup_{s\in S\cup\{*\}}\{\text{Know}_s\}$;
			\end{algorithmic}
        \end{algorithm}
        \end{minipage}
        \hfill
        \begin{minipage}[t]{0.48\textwidth}
  \begin{algorithm}[H]
		\caption{RCStandpoint}
			\label{RCStandpoint}
			\textbf{Input}: A defeasible standpoint knowledge base $\mathcal{K}$ in normal form and a (non-standpoint sharpening) defeasible standpoint formula $\psi$ in normal form.\\
			\mbox{\textbf{Output}: True if $\psi$ is in the rational closure of} 
            \mbox{$\mathcal{K}$, otherwise False.}\\[-2.6ex]
			\begin{algorithmic}[1]
				\IF{$\psi=\phi_1\wedge\phi_2$}
				\IF{RCStandpoint($\phi_1$)$\,=\,$True and\\ \ \ \ \  RCStandpoint($\phi_2$)$\,=\,$True}
				\STATE\textbf{return} True;
				\ELSE 
				\STATE\textbf{return} False;
				\ENDIF
				\ELSIF{$\psi=\Box_s\phi$}
				\FOR{$K\in \text{Know}_s$}
				\IF{RCProp($K,\phi$)=False}
				\STATE \textbf{return} False;
				\ENDIF
				\ENDFOR
				\STATE \textbf{return} True;
				\ELSIF{$\psi=\Diamond_s\phi$}
				\FOR{$K\in \text{Know}_s$}
				\IF{RCProp($K,\phi$)$\,=\,$True}
				\STATE \textbf{return} True;
				\ENDIF
				\ENDFOR
				\STATE \textbf{return} False;
				\ENDIF
		\end{algorithmic}
		\end{algorithm}
\end{minipage}
\end{figure}
		Intuitively, Algorithm \ref{RCStandpoint} examines the modal operator occurring at the front of a DRSL formula $\xi$ and based on this decides which KLM knowledge bases to consider when entailment-checking. If $\xi=\Box_s\psi$, then we check if $\psi$ is in the rational closure of every KLM-propositional knowledge base which coincides with the standpoint $s$. That is, we check it is entailed by rational closure in every knowledge base in $\text{Know}_s$. If $\xi=\Diamond_s\psi$ then we check if $\psi$ is entailed in \emph{some} rational closure of a KLM knowledge base in $\text{Know}_s$. In the case where $\xi$ has no modal operator we equivalently can check for $\Box_*\xi$. Using this, we are able to propose a reasonable definition of rational closure for DRSL, which can be seen as a natural extension of rational closure in the propositional case.
		
		\begin{definition}
			We say that $\phi$ is in the \textbf{rational closure} of $\mathcal{K}$, or write $\mathcal{K}\dentails_{RC} \phi$, if and only if \textnormal{RCStandpoint($\mathcal{K},\phi$)$\,=\,$True.}
		\end{definition}

      \begin{note}
    Algorithm \ref{RCStandpoint} is easily adaptable to other instances of algorithmically defined consequence operations used for propositional KLM. For example, lexicographic closure is another non-monotonic entailment operator introduced in the literature for the propositional case \cite{lehmann:lexicographicreason}. A more general algorithmic approach for defeasible entailments in the propositional KLM setting is introduced by Casini et al. \cite{casini:beyondratclosure}, including the case for lexicographic closure. In such a case, the above algorithm for rational closure in standpoints could be extended to use other similar algorithms for defeasible entailment, by calling on the lexicographic closure algorithm instead of \texttt{RCProp}, for example. Hence, the use case for the above algorithm is not necessarily restricted to the case of rational closure.
\end{note}

		We can also show $\dentails_{RC}$ as defined above is indeed non-monotonic, as opposed to ranked entailment. Consider the example below.
		
		\begin{example}\label{example:basicnonmonotonicDRSL}
			Consider the knowledge base, $\mathcal{K}=\{p\rightarrow b,b\twiddle f,s\preceq *\}$. If we compute StandpointSplit($\mathcal{K}$), since all the non-subsumption items in our knowledge base are universal, we obtain just one set in our splitting $K_s=\mathcal{K}-\{s\preceq *\}$. Then, since there are no exceptional antecedents
			 we are able to use classical reasoning to conclude $\mathcal{K}\dentails_{RC}\{p\twiddle f,\Box_s(p\twiddle f)\}$. However, if we consider the amended knowledge base, $\mathcal{K}'=\mathcal{K}\cup\{\Box_s(p\twiddle \neg f),t\preceq *\}$ we are able to show that $\mathcal{K}'\ndentails_{RC}\Box_s(p\twiddle f)$ and $\mathcal{K}'\ndentails_{RC} p\twiddle f$. Hence, $\dentails_{RC}$ is non-monotonic. It is also interesting to note that since $p\twiddle \neg f$ is only asserted for $s$ and not for other standpoints, we still have $\mathcal{K}'\dentails_{RC}\Box_t(p\twiddle f)$.
		\end{example}

\begin{note}\label{note:dontincludek*}
        In Algorithm \ref{alg:standpointsplit}, it can be seen that we intentionally exclude $K_*$ as a knowledge base when we consider our rational closure. This is due to the fact that we do not want to construct a ``basic universal'' standpoint as an individual precisification when we are querying over our logic. If we allow for $K_*$ to be constructed we may have $\mathcal{K}\ndentails_{RC}\Box_* \phi$ simply because $\phi$ is not in the rational closure of $K_*$. On the other hand, we may conclude $\mathcal{K}\dentails_{RC}\Diamond_* \phi$ simply because $\phi$ is in the rational closure of $K_*$. In both cases, this seems undesirable. In the first case, we may lose conclusions when we are reasoning about what finite agents agree upon, when these agreements are not the consequence of some universal specified rule. In the worst case, if $\mathcal{K}_*=\emptyset$, there is nothing non-trivial in the rational closure of $K_*$ and we are unable to entail anything non-trivial of the form $\Box_*\phi$, even if every known standpoint $s\in S-\{*\}$ agrees upon $\phi$. This would be the case in Example \ref{example:basicnonmonotonicDRSL}. In the second case, it feels erroneous to conclude $\Diamond_* \phi$ based on the fact that $\phi$ is in the rational closure of $K_*$, since intuitively $\Diamond_* \phi$ reads ``there is a standpoint which can be interpreted as entailing $\phi$''. However, if the only such standpoint is $*$, then this standpoint does not correspond to any considered agents viewpoint, and so $\Diamond_* \phi$ does not seem a reasonable conclusion. One downside to the exclusion of $K_*$ is that it means that \texttt{RCStandpoint} is not usable in the case where $\mathcal{K}$ only consists of formulas in the form $\Box_* \psi$. However, in this case $\mathcal{K}$ is equivalent to a propositional KLM knowledge base and so propositional rational closure can be used for  entailment-checking. It is also possible to enforce the inclusion of a knowledge base equal to $K_*$ by adding $\Diamond_*\top$ to  $\mathcal{K}$. Since $\top$ adds only a trivial element to $K_*$, $K_*^{\top}$ is equivalent to $K_*$.
    \end{note}

    Having defined and motivated the previous algorithm as a natural extension to rational closure, we now consider the complexity of using \texttt{StandpointSplit} and  \texttt{RCStandpoint} to check for entailment in DRSL. 

    \begin{lemma}\label{lemma:splitptime}
        \texttt{StandpointSplit} is computable in polynomial time.
    \end{lemma}

    \begin{theorem}\label{theorem:complexityrcstandpoint}
        \texttt{RCStandpoint} is computable in $\text{P}^{\text{NP}}$, and if the materialization of each defeasible entailment in the knowledge base and queried formula is a Horn clause then \texttt{RCStandpoint} is computable in polynomial time.
    \end{theorem}

    This theorem follows a general pattern in the standpoint logic literature, which shows in various cases that standpoint modal operators can be added, sometimes with additional restrictions, to a certain system of logic without increasing the complexity of the logic \cite{alvarez:stdptlogicfocase, alvarezrudolphstrass:standpointEL}. The result here comes from the fact that \texttt{RCProp} is in $\text{P}^{\text{NP}}$. Then, if our reduced DRSL knowledge base has length $k$, the \texttt{RCStandpoint} algorithm consists of computing \texttt{RCProp} at most $k$ times over a knowledge base of at most $k$ propositional KLM formulas.

\subsection{Semantic Characterisation for Rational Closure}    
    Besides the previous algorithm, we can equivalently characterize the rational closure for a knowledge base $\mathcal{K}$ using our previously defined semantics for DRSL. In particular, we can define a single ranked interpretation which is a model for a (non-standpoint sharpening) formula $\xi$ if and only if $\mathcal{K}\dentails_{RC} \xi$.
  
		\begin{definition}
			Let $\mathcal{K}$ be a knowledge base in normal form, then define the interpretation $M_{RC}^\mathcal{K}=(\Pi,\sigma,\gamma)$ as follows:
			\begin{itemize}
				\item $\Pi=\{\pi_s\mid s\in S-\{*\}\}\cup(\bigcup_{s\in S\cup\{*\}} \{\pi_s^\phi\mid \Diamond_s\phi \in\mathcal{K}\text{ for some }s\in S\})$.
				\item $\sigma(s)=\{\pi_t\mid t\preceq^+ s\}\cup \{\pi_t^\phi\mid t\preceq^+s\}$ for each $s\in S$. 
				\item For each $\pi_s$, we have $\gamma(\pi_s)=R^{K_s}_{RC}$ and for each $\pi_s^\phi$, we have $\gamma(\pi^\phi_s)=R^{K_s^\phi}_{RC}$.
			\end{itemize}
			where $K_s$ and $K_s^{\psi}$ are the sets as described in the standpoint splitting algorithm, and $R^{K}_{RC}$ is the ranking function representing the rational closure of the (propositional) KLM knowledge base $K$.
		\end{definition}

This is constructed with a similar motivation to \texttt{RCStandpoint}. Any standpoint $s\in \mathcal{S}$ must have some ``base'' precisification which is characterized by the KLM propositions hold in every precisification for $s$. This corresponds to the rational closure ranked interpretation for this set of propositions. Then, any proposition $\Diamond_s\phi$ characterizes a new precisification of $s$ where $\phi$ is taken into account. It is worth noting that, as expected, $M_{RC}^\mathcal{K}$ is a well-defined model of $\mathcal{K}$.

  \begin{lemma}\label{lemma:rcmodel}
      $M_{RC}^\mathcal{K}=(\Pi,\sigma,\gamma)$ is a model of $\mathcal{K}$.
  \end{lemma}

Through construction, it is also clear that the algorithmic and semantic characterization of rational closure follow the same principles. That is, we split the standpoint knowledge base into various KLM-propositional knowledge bases, and connect these to a certain standpoint. Then, for each knowledge base, we perform a "pointwise" rational closure. In the algorithmic case, we call on the \texttt{RCProp} algorithm, and use the ranked interpretations for rational closure in the semantic case. From this, we are able to see that entailment checking under the rational closure model $M_{RC}^\mathcal{K}$ gives the same result as the algorithm \texttt{RCStandpoint}. This is formalized with the following theorem. 

\begin{theorem}\label{theorem:modelalgorithmrepresentation}
    For all $\psi\in \mathcal{L}^{\twiddle}_{\mathds{S}}$ and all $\mathcal{K}\subseteq \mathcal{L}^{\twiddle}_{\mathds{S}}$, we have that $\mathcal{K}\dentails_{RC}\psi$ if and only if $M_{RC}^\mathcal{K}\Vdash \psi$.
\end{theorem}

This gives us the main result of our paper. As in the propositional case for KLM, we have a non-monotonic entailment operation which can be equivalently characterized through an algorithm, and a single representative semantic model. We now consider applying the above to our original motivating example. 

\begin{example}\label{example:finaltomatoes}
    Using the same knowledge base as Example \ref{example:originaltomatoes} we use the splitting defined in Example \ref{example:tomatosplit} to construct our ranked interpretations at each precisification, where $\gamma(\pi_B)$, $\gamma(\pi_C)$ and $\gamma(\pi_L)$ are the ranked interpretations defined by the rational closure of $K_B$, $K_C$ and $K_L$, respectively. Since $K_B$ is a Boolean knowledge base, the rational closure contains every valuation satisfying $K_B$ in row $0$ and every other valuation in row $\infty$. For the other two knowledge bases we have the following ranked interpretations corresponding to rational closure. In the representation of these interpretations, we change each atomic proposition to the first letter of its name (eg. $t=tomato$) and the propositions $sweet$ and $savory$ are represented by $s_w$ and $s_a$ respectively:
   \[ \begin{tabular}
    {|c|c|c|}
      \hline  
      $\infty$ & all other valuations & all other valuations  \\
      \hline
        1 & $ts_a vs_wf$, $t\bar{s_a v}s_wf$, $t\bar{s_a vs_wf}$,$\bar{t}s_a vs_wf$ &  $t\bar{s_a v}s_wf$, $t\bar{s_a vs_wf}$\\
        \hline
        0 & $ts_a v\bar{s_wf}$, $\bar{t}s_a v\bar{s_wf}$,$\bar{ts_a v}s_wf$, $\bar{ts_a vs_wf}$ & $ts_a v\bar{s_wf}$, $\bar{t}s_a v\bar{s_wf}$,$\bar{ts_a v}s_wf$, $\bar{ts_a vs_wf}$\\
        \hline
    \end{tabular}\]
    Here, the middle column represents the ranking function of $\gamma(\pi_C)$ and the rightmost represents $\gamma(\pi_L)$. We are then able to use these models to check if DRSL statements are in the rational closure for $\mathcal{K}$. First, note that $\gamma(\pi_C)\Vdash tomato\twiddle vegetable$ and $\gamma(\pi_L)\Vdash tomato\twiddle vegetable$. Moreover, $\gamma(\pi_B)\Vdash tomato \rightarrow vegetable$, which is a stronger condition than $tomato\twiddle vegetable$. Thus, $M^\mathcal{K}_{RC},\pi\Vdash tomato\twiddle vegetable$ for each $\pi\in \Pi$ and $\mathcal{K}\dentails_{RC} tomato\twiddle vegetable$. We are also able to check for non-entailment. For example, $\mathcal{K}\ndentails_{RC} \Box_L(tomato\rightarrow \neg fruit)$ since the valuation $t\bar{s_a v}s_wf$ has rank 1 under $\gamma(\pi_L)$. This seems reasonable intuitively, since it allows for the legal position to consider a tomato a fruit in an exceptional case, such as the possibility of a sweet "dessert tomato".
\end{example}

\begin{note}
    It is worth noting here that we have been operating under the assumption that $S\cup\{*\}$, the set of named standpoints in the knowledge base, is equal to $\mathcal{S}$, the set of all standpoints expressible in our vocabulary. This is akin to a closed-world assumption with regards to standpoints in the given knowledge base. This may seem restrictive since we are unable to query for a standpoint which is not named in the knowledge base. However, the addition of standpoints which have no formulas in the knowledge face a similar problem to the addition of $K_*$ as in Note \ref{note:dontincludek*}. For example, if we added another standpoint $X$ to Example \ref{example:finaltomatoes} then we are required in our model to consider $\pi_X$. However, since there are no statements pertaining to $X$, $\gamma(\pi_X)(u)=0$ for any $u\in\mathcal{U}$. Then we are no longer able to conclude $tomato\twiddle vegetable$\footnote{In fact, we lose entailment for any non-trivial statements of the form $\Box_*\phi$.} in our model simply due to the fact that it does not hold in $\gamma(\pi_X)$. Therefore, we keep this assumption, and note that in the case where we wish to consider an additional viewpoint $X$ that does not already occur in the knowledge base, we can equivalently add the statement $X\preceq *$ into the knowledge base as in Example \ref{example:basicnonmonotonicDRSL}.
\end{note}

\section{Related Work}

An early notion of a logical semantics incorporating standpoints was introduced by Bennet \cite{bennett2011standpoint}, using a fairly different approach to those used in this paper. The syntax and semantics for standpoint modalities used in this paper was developed in the propositional case by Gómez Álvarez and Rudolph \cite{alvarezrudolph:propositionalstdpt}. Similar notions of standpoints have since been introduced to first order logic \cite{alvarez:stdptlogicfocase}, various description logics \cite{alvarez:stdptlogicfocase,alvarezrudolphstrass:standpointEL,alvarezstrassrudolph:standpointELplus} and linear temporal logic \cite{alvarezlyon:stndpttemporal}. The defeasible logic of KLM was originally introduced by Kraus et al. \cite{kraus:nonmonotonic} with rational closure defined by Lehmann and Magidor \cite{lehmann:conditionalentail}. Although defeasible implication has not been studied previously in the case for standpoints, similar notions have been previously considered. Britz et al. \cite{britzmeyervar:normalmodalpreferential, britzetal:preferentialreasoningmodal} consider introducing $\twiddle$ as a non-monotonic consequence relation between propositions in modal logics. Another aspect of defeasibility in modal logic are \textit{defeasible modal operators}, which have been considered by Britz and Varzinczak \cite{britzvarzin:defeasiblemodalities} in the general case and Chafik et al. \cite{chafik:defeasiblelineartemporal} in the case of linear temporal logic. 
 Other non-modal extensions of KLM-style defeasibility and derived non-monotonic entailments have been considered in the description logics case \cite{britz2020principles, casini2012lexicographic, casini2014relevant}, as well as for first order logic and some of its decidable fragments\cite{delgrande2020preference,meyerguyPJ:klmrfol,MorrisEtAl:disjunctdatalog}.

\section{Conclusion}

The main focus of this paper was to integrate standpoint modal operators into KLM-style defeasible propositional logic. In order to do this we defined Defeasible Restricted Standpoint Logic (DRSL), by introducing a syntax which allows KLM propositional symbols to be bound by standpoint modal operators. We introduced the semantics of DRSL through ranked standpoint structures, combining the notion of ranked interpretations for propositional KLM \cite{lehmann:conditionalentail}, and the semantic approaches used to define other standpoint logics \cite{alvarezrudolph:propositionalstdpt,alvarez:stdptlogicfocase}. We then introduced a non-monotonic entailment relation for DRSL, which we proposed as a natural extension to rational closure in the propositional case \cite{lehmann:conditionalentail}, denoted $\dentails_{RC}$. This was first defined via a rational closure algorithm, which we showed provided a method for entailment-checking in DRSL that was of the same complexity class as the rational closure algorithm in the propositional case. Lastly, we showed that $\dentails_{RC}$ could be equivalently defined semantically, specifically constructing a single ranked standpoint structure $M^\mathcal{K}_{RC}$ such that $\mathcal{K}\dentails_{RC}\psi$ if and only if $M^\mathcal{K}_{RC}\Vdash \psi$. 

For future work, the most obvious question would be to enquire whether a more expressive version of DRSL can be defined (for example including negation and disjunction of standpoint bound formulas), and what effects allowing for a more expressive logic would have on the complexity and semantic constructions linked to a non-monotonic entailment, such as rational closure. It seems initially that this case would be significantly more difficult to consider, since we lose the Single Model property when we introduce negation and disjunction into KLM in the propositional case \cite{meyerPJ:BKLM}. Another notion to explore, although mentioned briefly in the paper, is the possibility of extending other non-monotonic consequence relations defined in the propositional case to the standpoint case, such as lexicographic closure \cite{lehmann:lexicographicreason}. It would also be desirable to investigate the proof-theoretic properties of $\dentails_{RC}$ in the standpoint logic case and compare those to the original KLM postulates given by Kraus et al. \cite{kraus:nonmonotonic}. Another interesting problem to investigate would be the integration of standpoints into more expressive KLM-style defeasible logics, such as defeasible description logics (DDLs) \cite{britz2020principles}. Since standpoint have seen to be integratable in other description logics \cite{alvarez:stdptlogicfocase,alvarezrudolphstrass:standpointEL}, it seems plausible that this would work in the case of DDLs.

\begin{credits}
\subsubsection{\ackname} 
Nicholas Leisegang wishes to thank the School of Embedded Composite Artificial Intelligence (SECAI) -- project 57616814 funded by BMBF (the Bundesministerium für Bildung und Forschung) and DAAD (German Academic Exchange Service) -- 
and the University of Cape Town's Artificial Intelligence Research Unit (AIRU), who facilitated the research visit to TU Dresden which made this collaboration possible. Nicholas Leisegang is funded by the UCT Science Faculty Fellowship.

\subsubsection{\discintname}The authors have no competing interests to declare that are
relevant to the content of this article.
\end{credits}

\clearpage

\appendix
\section{Appendix}
\subsection{Proofs of Results in Sections \ref{section:DRSLsyntaxsemantic} and \ref{section:DRSLrationalclosure}}

\textbf{Lemma \ref{lemma:boxdistribution}.} \textit{For any ranked standpoint structure $M$ and any $\psi_1,\psi_2\in \mathcal{L}^{\twiddle}_{\mathds{S}}$, we have that $M\Vdash \Box_s(\psi_1\wedge \psi_2)$ if and only if $M\Vdash \Box_s\psi_2\wedge \Box_s\psi_2$.}

\begin{proof}
Consider two formulas $\psi_1,\psi_2\in\mathcal{L}^{\twiddle}_{\mathds{S}}$. Then,

$M\Vdash \Box_s(\psi_1\wedge \psi_2)$ 

$\iff M,\pi\Vdash \psi_1\wedge \psi_2$ for all $\pi\in\sigma(s)$,

$\iff M,\pi\Vdash \psi_1$ and $M,\pi\Vdash \psi_2$ for all $\pi\in\sigma(s)$,

$\iff M\Vdash\Box_s \psi_1$ and $M\Vdash\Box_s \psi_2$,

$\iff M\Vdash \Box_s\psi_2\wedge \Box_s\psi_2$.

\end{proof}
\textbf{Lemma \ref{lemma:diamonddistribution}.} 
\textit{Let $M$ be a ranked standpoint structure, $\phi_1,...,\phi_k\in\mathcal{L}^{\twiddle}$. Then $M\Vdash \Diamond_s(\phi_1\wedge\#^2_{s_2}\phi_2\wedge ...\wedge\#^k_{s_k} \phi_k)$ if and only if $M\Vdash\Diamond_s\phi_1\wedge\#_{s_2}\phi_2\wedge ...\wedge\#_{s_k} \phi_k$.}
\begin{proof}
    $(\Rightarrow):$ Assume $M\Vdash \Diamond_s(\phi_1\wedge\#^2_{s_2}\phi_2\wedge ...\wedge\#^k_{s_k} \phi_k)$. Then, $M,\pi\Vdash \phi_1\wedge\#^2_{s_2}\phi_2\wedge ...\wedge\#^k_{s_k} \phi_k$ for some $\pi\in \sigma(s)$. This implies $M,\pi\Vdash \phi_1$ and $M,\pi\Vdash \#^i_{s_i} \phi_i$ for each $i\in\{2,..,k\}$, which is equivalent to $M\Vdash \Diamond_s\phi_1$ and $M,\Vdash \Diamond_s\#^i_{s_i} \phi_i$ for each $i\in\{2,..,k\}$. Then, since multiple modal operators reduce to the rightmost one by a previous result, this is equivalent to $M\Vdash \Diamond_s\phi_1$ and $M\Vdash \#^i_{s_i} \phi_i$ for each $i\in\{2,..,k\}$. Hence, $M\Vdash\Diamond_s\phi_1\wedge\#^2_{s_2}\phi_2\wedge ...\wedge\#^k_{s_k} \phi_k$.

    $(\Leftarrow):$ Assume $M\Vdash\Diamond_s\phi_1\wedge\#^2_{s_2}\phi_2\wedge ...\wedge\#^k_{s_k} \phi_k$. Then $M\Vdash \Diamond_s\phi_1$ and $M,\Vdash \#^i_{s_i} \phi_i$ for each $i\in\{2,..,k\}$. $M\Vdash \Diamond_s\phi_1$ is equivalent to $M,\pi\Vdash \phi_1$ for some $\pi\in \sigma(s)$. Moreover, for the same $\pi \in\sigma(s)$, we have that $M,\pi \Vdash\#^i_{s_i} \phi_i$ for each $i\in\{2,..,k\}$ since each $\#^i_{s_i}\phi_i$ is entailed globally. Therefore  $M,\pi\Vdash \phi_1\wedge\#^2_{s_2}\phi_2\wedge ...\wedge\#^k_{s_k} \phi_k$ and so $M\Vdash \Diamond_s(\phi_1\wedge\#^2_{s_2}\phi_2\wedge ...\wedge\#^k_{s_k} \phi_k)$.
\end{proof}
\textbf{Corollary \ref{corollary:propositionnormalform}.} \textit{Any DRSL formula $\psi$ can be represented equivalently by a formula in the form $\#^1_{s_1}\phi_1\wedge...\wedge \#^k{s_k}\phi_k$, where $\phi_1,...,\phi_n\in\mathcal{L}^{\twiddle}$.}

\begin{proof}
    We begin in the ``basic cases'' where our formula $\psi$ is equal to $\phi$ or equal to $\#_s\phi$ for a single standpoint modal operator $\#_s$ and some $\phi \in\mathcal{L}^{\twiddle}$. In the second case $\psi$ is already in the desired form and if $\psi\in \mathcal{L}^{\twiddle}$ we have by a previous result that $\psi$ is equivalent to $\Box_*\psi$ and it is then in the desired form.

    If we take a formula in the above form and add an arbitrary (finite) number and combination of modal operators in front of it $\#^1_{s_1}...\#^k_{s_k}$ we have that if $\psi=\phi$ then $\#^1_{s_1}...\#^k_{s_k}\phi$ is equivalent to $\#^k_{s_k}\phi$. If $\psi=\#_s\phi$ then $\#^1_{s_1}...\#^k_{s_k}\psi$ is equivalent to $\psi$. It is also clear to see that if we take any finite number of conjunctions of ``basic'' formulas, we obtain a formula still in normal form.

    Finally, we consider a case where we apply a standpoint operator to a finite conjunction of basic formulas. From the previous reasoning, and the commutativity of $\wedge$ we can without loss of generality express this conjunction in the form $\phi_1\wedge\#^2_{s_2}\phi_2\wedge ...\wedge\#^k_{s_k} \phi_k$ for $\phi_i\in\mathcal{L}^{\twiddle}$. Then, by Lemma \ref{lemma:boxdistribution}, $\Box_s(\phi_1\wedge\#^2_{s_2}\phi_2\wedge ...\wedge\#^k_{s_k} \phi_k)$  is equivalent to $\Box_s\phi_1\wedge\#^2_{s_2}\phi_2\wedge ...\wedge\#^k_{s_k} \phi_k$ and by Lemma \ref{lemma:diamonddistribution}, $\Diamond_s(\phi_1\wedge\#^2_{s_2}\phi_2\wedge ...\wedge\#^k_{s_k} \phi_k)$  is equivalent to $\Diamond_s\phi_1\wedge\#^2_{s_2}\phi_2\wedge ...\wedge\#^k_{s_k} \phi_k$ and we are done.
\end{proof}
\textbf{Lemma \ref{lemma:rankedentailmentmonotonic}}. \textit{Let $\mathcal{K}$ be a DRSL knowledge base. Then, for any $\psi\in \mathcal{L}^{\twiddle}_{\mathds{S}}$, $\mathcal{K}\vDash_R \xi$ implies that $\mathcal{K}\cup\{\psi\}\vDash_R \xi$.}

\begin{proof}
    Assume $\mathcal{K}\vDash_R \xi$. By definition $M\Vdash \xi$ for any model $M$ of $\mathcal{K}$. Then, in particular any model $M'$ of $\mathcal{K}\cup\{\psi\}$ is a model of $\mathcal{K}$. Hence, $M'\Vdash \xi$.
\end{proof}
\textbf{Lemma \ref{lemma:splitptime}.} \textit{\texttt{StandpointSplit} is computable in polynomial time.}

\begin{proof}
    Let $k$ be the size of the knowledge base $\mathcal{K}$. Then note that since the only standpoints we consider are those named in the knowledge base, there are at most $k$ elements in $S$. Hence we only complete the for loop in line 2 at most $k$ times. The same applies to the while and for loops in lines 4 and 5. Thus without considering the complexity of checking whether $s\preceq^+ t$, this loop takes $O(k^3)$ steps. Then checking for each $t\in S$, if $s\preceq^+ t$ for some fixed $s\in S$ (that is, computing the transitive closure of $\preceq$) is known to be in polynomial time.

    The number of times repeating the loops in lines 10 and 11 are again bounded by the size of $\mathcal{K}$ and so the second loop of the algorithm takes $O(k^2)$ steps. The repetitions last for loop in line 15 is again bound by the size of $\mathcal{K}$, and checking which elements are in $S_{\preceq s}$ is another instance of transitive closure checking, and since the knowledge bases are already identified and indexed, computing $\text{Know}_s$ is trivial. Hence, each loop is computable in polynomial time. 
\end{proof}
\textbf{Theorem \ref{theorem:complexityrcstandpoint}.} \textit{\texttt{RCStandpoint} is computable in $\text{P}^{\text{NP}}$ time, and if the materialization of each defeasible entailment in the knowledge base and queried formula is a Horn clause then \text{RCStandpoint} is computable in P time.
}

\begin{proof}
    We start by considering the case where $\psi$, the formula being queried, is of the form $\Box_s\phi$ or $\Diamond_s \phi$ for standpoint symbol $s\in S$ and $\phi\in\mathcal{L}^{\twiddle}$. Let $k$ be the size of the knowledge base and $n$ be the size of the query $\psi$. That is, $n$ is the number of ``basic'' defeasible implications in $\psi$.

    In the worst case, each element in $\mathcal{K}$ generates a new KLM knowledge base in $\text{Know}_s$ and so we repeat the loop at line 8 or 15 at most $k$ times. Furthermore, at worst the KLM knowledge base $K$ is the size of of the original knowledge base $\mathcal{K}$. It is shown by Lehman and Magidor \cite{lehmann:conditionalentail} that checking whether a single defeasible implication $\alpha\twiddle\beta$ is in the rational closure of $K$ requires at most $O(k^2)$ Boolean satisfiability checks. Since we are checking at most $n$ defeasible implications we require at most $O(nk^2)$ checks. Then the whole loop is computable with $O(nk^3)$ Boolean satisfiability checks (where we use our NP oracle). Hence, in this case \texttt{RCStandpoint} is in $\text{P}^{NP}$.

    If $\psi$ is a conjunction of formulas then at worst we repeat the \texttt{RCStandpoint} once for each conjunct. That is, at worst $n$ times. Therefore, in this case we are still in $\text{P}^{\text{NP}}$. Furthermore, since \texttt{StandpointSplit} is in polynomial time the problem of identifying the members in $\text{Know}_s$ does not increase our complexity class.

    Now we consider the case where we take the materialization of each formula in each KLM knowledge base $K\in\text{Know}_*$ as well as the materialization of the subformulas of $\psi$ in $\mathcal{L}^{\twiddle}$. That is, we consider these formulas and replace every instance of ``$\twiddle$'' with ``$\rightarrow$''. If each formula considered is Horn, then, as in the propositional case \cite{lehmann:conditionalentail} every Boolean satisfiability check is computable in polynomial time and therefore \texttt{RCStandpoint} is computable in polynomial time.
\end{proof}
\textbf{Lemma \ref{lemma:rcmodel}.} \textit{$M_{RC}^\mathcal{K}=(\Pi,\sigma,\gamma)$ is a model of $\mathcal{K}$.}

\begin{proof}
    For any standpoint $s\in S\cup\{*\}$, it is clear to see there is a bijection between $\text{Know}_s$ and $\sigma(s)$ defined by $K_t\mapsto \pi_t$ and $K_t^\phi \mapsto \pi_t^\phi$. Then, for any $K\in \text{Know}_s$, we have that there is a unique $\pi\in \sigma(s)$ such that $\gamma(\pi)=R^{K}_{RC}$.

    Since $\mathcal{K}$ is in normal form we only have to consider two cases:
    \begin{itemize}
        \item If $\psi\in\mathcal{K}$ is of the form $\Box_s \phi$ where $s\in\mathcal{S}$ and $\phi\in\mathcal{L}^{\twiddle}$, then note that $\phi\in K$ for all $K\in\text{Know}_s$. Then for each $\pi\in \sigma(s)$, we have by our bijection that $\gamma(\pi)=R^{K}_{RC}$ for some $K\in\text{Know}_s$. Hence $\gamma(\pi)=R^{K}_{RC}\Vdash \phi$ by definition and so $M,\pi\Vdash \phi$ for all $\pi\in\sigma(s)$. Hence, $M\Vdash \Box_s \phi$.
        \item If $\psi\in\mathcal{K}$ is of the form $\Diamond_s \phi$ where $s\in\mathcal{S}$ and $\phi\in\mathcal{L}^{\twiddle}$, then in particular $\phi\in K_s^{\phi}$. Then $R^{K_s^\phi}_{RC}\Vdash \phi$ by definition and $R^{K_s^\phi}_{RC}=\gamma(\pi_s^{\phi})$. Lastly note that $\pi_s^{\phi}\in\sigma(s)$ and so $M\Vdash \Diamond_s \phi$.
    \end{itemize}
\end{proof}   
\textbf{Theorem \ref{theorem:modelalgorithmrepresentation}.}   \textit{For all $\psi\in \mathcal{L}^{\twiddle}_{\mathds{S}}$ and all $\mathcal{K}\subseteq \mathcal{L}^{\twiddle}_{\mathds{S}}$, we have that $\mathcal{K}\dentails_{RC}\psi$ if and only if $M_{RC}^\mathcal{K}\Vdash \psi$.}

\begin{proof}

 Let $s$ be a standpoint symbol and $\phi\in\mathcal{L}^{\twiddle}$. If $\psi=\Diamond_s\chi$, then (using $M$ as shorthand for $M^\mathcal{K}_{RC}$) the following are equivalent:
 \begin{itemize}
     \item RCStandpoint($\mathcal{K},\psi$)=True.
     \item RCProp($K,\chi$)=True for some $K\in\text{Know}_s$.
     \item $R^{K}_{RC}\Vdash \chi$ for some $K\in\text{Know}_s$. 
     \item $\gamma(\pi)\Vdash \chi$ for some $\pi\in\sigma(s)$ .
     \item $M,\pi\Vdash \chi$ for some $\pi\in\sigma(s)$.
     \item $M\Vdash \Diamond_s\chi$.
 \end{itemize}

  If $\psi=\Box_s\chi$, then the following are equivalent:
  \begin{itemize}
     \item RCStandpoint($\mathcal{K},\psi$)=True.
     \item RCProp($K,\chi$)=True for all $K\in\text{Know}_s$.
     \item $R^{K}_{RC}\Vdash \chi$ for all $K\in\text{Know}_s$. 
     \item $\gamma(\pi)\Vdash \chi$ for all $\pi\in\sigma(s)$.
     \item $M,\pi\Vdash \chi$ for all $\pi\in\sigma(s)$.
     \item $M\Vdash \Box_s\chi$.
  \end{itemize}

Note that in each case, we use the existence of a bijection between $\text{Know}_s$ and $\sigma(s)$ in the step from the third to fourth point. We also use the representation theorem for the KLM-propositional case for the second to third point. 

Then if $\psi$ is a formula in $\mathcal{L}^{\twiddle}$, it is equivalent to $\Box_* \psi$ and thus is solved by the previous case.  Lastly, if $\psi$ is a conjunction of formulas $\chi_1,...,\chi_k$ from our previous cases, then $\mathcal{K}\dentails_{RC} \psi$  iff $\mathcal{K}\dentails_{RC} \chi_i$ for each $\chi_i$. This is true iff $M\Vdash \chi_i$ for all $\chi_i$ which again is equivalent to $M\Vdash \psi$. 
\end{proof}

\end{document}